\documentclass[graybox]{svmult}
\usepackage{mathptmx}       % selects Times Roman as basic font
\usepackage{helvet}         % selects Helvetica as sans-serif font
\usepackage{courier}        % selects Courier as typewriter font
\usepackage{type1cm}        % activate if the above 3 fonts are
\usepackage{makeidx}         % allows index generation
\usepackage{graphicx}        % standard LaTeX graphics tool
\usepackage{multicol}        % used for the two-column index
\usepackage{CJKutf8}
\usepackage[bottom]{footmisc}% places footnotes at page bottom

\makeindex             % used for the subject index
\begin{document}

\title*{Can Large Language Models be Used to Provide Psychological Counselling? \\
An Analysis of GPT-4-Generated Responses Using Role-play Dialogues}
\titlerunning{Can Large Language Models Be Used to Provide Psychological Counselling?}

\author{Michimasa Inaba, Mariko Ukiyo and Keiko Takamizo}

\institute{Michimasa Inaba \at The University of Electro-Communications, Chofu, Tokyo, Japan. \email{m-inaba@uec.ac.jp}
\and  Mariko Ukiyo\at iDEAR Human Support Service, Shibuya, Tokyo, Japan.
\and Keiko Takamizor \at Japanese Organization of Mental Health and Educational Agencies, Shibuya, Tokyo, Japan. \email{takamizo@idear.co.jp}}

\maketitle

\abstract*{}

\abstract{Mental health care poses an increasingly serious challenge to modern societies. In this context, there has been a surge in research that utilizes information technologies to address mental health problems, including those aiming to develop counseling dialogue systems. However, there is a need for more evaluations of the performance of counseling dialogue systems that use large language models. For this study, we collected counseling dialogue data via role-playing scenarios involving expert counselors, and the utterances were annotated with the intentions of the counselors. To determine the feasibility of a dialogue system in real-world counseling scenarios, third-party counselors evaluated the appropriateness of responses from human counselors and those generated by GPT-4 in identical contexts in role-play dialogue data. Analysis of the evaluation results showed that the responses generated by GPT-4 were competitive with those of human counselors.}

\section{Introduction}

Mental health care poses an increasingly serious challenge for modern societies. For example, in Japan, suicide is the leading cause of death among those aged 10 to 39 \cite{WhitePaper}, and the World Health Organization has reported that suicide is a major cause of death among young people globally \cite{who}.
Consequently, text-based counselling, which refers to providing psychological support using instant messaging apps, has become increasingly prominent in recent years. Text-based counselling is more accessible than telecounselling and e-mail-based counseling, especially for the younger generation. However, the shortage of counselors proﬁcient in text-based counseling is a major obstacle. Even those who are experienced in in-person and e-mail-based counseling, as well as telecounselling, encounter difﬁculties in providing text-based counseling without appropriate guidance and training. Moreover, there is a shortage of personnel who can provide such guidance. 

Researchers are currently studying the uses of natural language processing for providing mental health support. Monitoring is an important aspect of mental health management, and the automatic detection of mental health disorders represents a particularly active area of current research \cite{gkotsis2017characterisation, chancellor2020methods, ji2018supervised, nguyen-etal-2022-improving, saha2022mental}.

In the field of dialogue systems, several systems aiming to improve mental health have been developed. 
For example, a system known as Woebot \cite{fitzpatrick2017delivering} facilitates cognitive behavioral therapy. There is also a system that provides emotional support \cite{cheng-etal-2023-pal,liu-etal-2021-towards}, and a system that responds by rewriting less empathetic utterances as more empathetic statements \cite{towards2021}.

Recent developments with large language models (LLM) have demonstrated their adaptability to various tasks and domains. However, the performance of counselling dialogue systems that utilize LLMs has not been fully evaluated. 
The ChatCounselor \cite{liu2023chatcounselor}  system is the most similar to the system used in our study because this LLM-based model was constructed to answer clients' questions and requests as a counsellor. 
However, ChatCounselor only supports single-turn question-answering and does not support multiturn dialogue. Furthermore, in the study of ChatCounselor, the generated responses were evaluated automatically by GPT-4 and not by expert counsellors.

In this study, we constructed a counselling dialogue system using GPT-4, and professional counsellors evaluated the generated responses. 
To generate appropriate responses, we collected counselling dialogue data via role-playing scenarios involving expert counsellors, and the utterances were annotated with the intentions of the counsellors. To assess the feasibility of using a dialogue system in actual counselling situations, third-party counsellors evaluated the appropriateness of the responses given by the human counsellors and those generated by the GPT-4 during identical contexts in the role-play dialogue data.

\section{The collection of role play dialogue}

In order to collect data for analysis, two counsellors participated in role-playing scenarios, with one counsellor playing the client and the other playing the counsellor. 
The dialogues were conducted in Japanese using the messaging application LINE, and data were collected for a total of six dialogues, one for each of the six themes shown in Table \ref{tb:theme}. 
Table \ref{tb:stats} shows the number of words and utterances, and Table \ref{tb:dlg_ex_theme1} shows a part of the collected role-play dialogues for Theme 1.

\begin{table}
\centering
\caption{List of role-play dialogue themes}
    \begin{tabular}{p{0.3cm}p{1.0cm}p{9.9cm}} \hline
    & 1 & (\begin{CJK}{UTF8}{ipxm}本番になると緊張してしまいます、どうしたらいいですか？\end{CJK} (I get nervous at competitions. What should I do?) \\ 
    & 2 & \begin{CJK}{UTF8}{ipxm}パートナーのことは好きだけど束縛が強くてつらい\end{CJK} (I love my partner, but it's hard because she is so restrictive.)\\ 
    & 3 & \begin{CJK}{UTF8}{ipxm}友だちとケンカしてしまいました、どうしたらいいですか？\end{CJK} (I got into a fight with a friend. What should I do?)\\ 
    & 4 & \begin{CJK}{UTF8}{ipxm}勉強する気がありません、どうしたらいいですか？\end{CJK} (I don't want to study. What should I do?) \\ 
    & 5 & \begin{CJK}{UTF8}{ipxm}男とか女とか決めつけられたくないです\end{CJK} (I don't want to be categorized as male or female.) \\ 
    & 6 & \begin{CJK}{UTF8}{ipxm}生きる意味ってなんだろう？\end{CJK} (What is the meaning of life?)\\ \hline
    \end{tabular}
    
    \label{tb:theme}
\end{table}

\begin{table}[!htb]
    \centering
    \caption{The number of words and utterances from six role-play dialogues}
    \begin{tabular}{p{3cm}p{2cm}p{2cm}} \hline
    & Utterances & Words \\ \hline
    Client & 1,194 & 15,962 \\ 
    Counsellor & 1,140 & 31,689 \\ 
    GPT-4 generated & 816 & 43,833 \\ \hline
    \end{tabular}
    
    \label{tb:stats}
\end{table}

\begin{table*}[!htb]
    \centering
    \caption{Example of role-play dialogue and counsellors' key points and intent annotations for Theme
    1. The abbreviation CLI stands for client, and COU stands for counsellor. The dialogues were conducted originally in Japanese and then translated into English by the authors.
}
    \begin{tabular}{l p{4.3cm}|p{2.7cm}|p{3cm}}
        Utterance &  & Key point & Intent \\ \hline \hline
        CLI: & I am a gymnast. Right before a competition, I get nervous and don't know what to do. & ~ & ~ \\ \hline
        COU: & I see.  & Before expressing an opinion or asking a question, first offer an affirmative answer. & To show that the counsellor is listening to the client.  \\ \hline
        COU: & Sometimes when you're nervous, you feel different than usual.  & ~ & To convey that it is nothing special.  \\ \hline
        ~ & ...  & ~ & ~ \\ \hline
        CLI: & By sorting out what things make you nervous, you may be able to find a way to cope. & Encourage concretization of the cause of the nervousness. & Because the content touches on personal values, tell the client what is generally said to make it easier for them to accept. \\ \hline
        CLI: & For what? Not because it's a competition? & ~ & ~ \\ \hline
        COU: & For example, you may be nervous because everyone is watching you at a competition and you may wonder what they think of you,   & Because the text is long, send it once and send the rest of the text later. & To provide concrete examples and to make it easier to imagine.  \\ \hline
    \end{tabular}
    \label{tb:dlg_ex_theme1}
    
\end{table*}

\section{Counsellor response generation by GPT-4}

To analyze whether an LLM can successfully engage in counselling dialogues, we used the role-play dialogue data to prompt GPT-4 to generate utterances while assuming the role of a counsellor.

In LLM prompts, chain-of-thought (CoT) prompting \cite{wei2022chain} has been found to be effective at not only generating answers to questions and problems but also facilitating the process of thinking about them. 
To perform CoT prompting and generate higher-quality responses, we annotated the collected counsellor utterances with the response key points and the intent of the response (Table \ref{tb:dlg_ex_theme1}). For themes one through three, the annotations were performed by the counsellor-role speakers; for themes four through six, this was done by counsellors who did not participate in the role-play dialogue.

\begin{table}[!htb]
    \centering
    % \small
    \caption{The prompt provided to GPT-4 for generation. To facilitate a process of thinking similar to chain-of-thought prompting, the counselor's key points and intentions annotated in the role-play data (the texts enclosed in ``[ ]") were added before the counselor's utterances in the dialogue section. Sentences highlighted in green are GPT-4-generated intentions and responses. The prompt was originally written in Japanese and translated by the authors.}
    \begin{tabular}{|p{11.5cm}|} \hline
\# Task \\
You provide text-based counseling. As a counselor, you should generate appropriate responses for your client and the intentions behind them. Intentions should be written in [ ] before the response.  \\ \\
\# Counseling Guidelines \\
$\cdot$ Show empathy to close distance with the client \\
$\cdot$ Before expressing an opinion or asking a question, provide an agreeable response first \\
$\cdot$ After offering an opinion, check how the client feels about it \\
$\cdot$ Summarize the client's concerns and thoughts, and conﬁrm that what the counselor understands is what the client wanted to talk about \\
$\cdot$ Communicate a positive perspective that the client may not be aware of \\
$\cdot$ Provide concrete examples to help the client visualize what is happening\\ \\

\# Counselling Dialogue \\
Counsellor: [Greetings] Good morning. \\
Client: Good morning 	\\
Counsellor: [Expresses gratitude for the client's decision to come for counselling. The client may have had the courage to come to you for counselling, and 
this is to put the client at ease.] Thank you for coming for counselling today. \\
Counsellor: [Asks broadly at first.] Please tell me what you would like to talk about today. \\
Client: OK \\
Client: Can I start now? \\
Counsellor:\colorbox[rgb]{0.8, 1.0, 0.8}{[Reassure the client that this is an environment where they could feel comfortable }  
\colorbox[rgb]{0.8, 1.0, 0.8}{to talk.] Sure. You can talk about whatever you want. Please feel free to do so.} \\ \hline
    \end{tabular}
    \label{tb:prompt}
\end{table}

The prompt we gave to GPT-4 for response generation is shown in Table \ref{tb:prompt}.
The prompt began with instructions telling GPT-4 to generate responses as a counsellor. Moreover, counsellor guidelines that were developed under the supervision of professional counsellors were also provided to GPT-4. Next, we included the ongoing dialogue between the counsellor and the client in the prompts. The provided dialogue included all text from the start of the dialogue up until the previous client's utterance. In addition, the annotated key points and intentions were added before the counsellor's utterance in the dialogue, and the LLMs were instructed to generate them as well. Table 5 shows the generation examples, and the texts enclosed in ``[ ]" are the generated intent.

We used GPT-4-0613 via OpenAI API, with a temperature of 0.0 and default values for the other parameters. Numerical details of the generated utterances are displayed in Table \ref{tb:stats}. 
As shown, the number of human counsellor utterances is higher than the number of utterances generated by GPT-4 because, in the role-play scenario, human speakers were able to send multiple consecutive messages. In contrast, GPT-4 generated one utterance at a time, as shown in Table \ref{tb:dlg_gpt}.

\begin{table}
\caption{Examples of intent and response generation results by GPT-4. The score represents the average of the three professional counsellors' evaluation scores for each generated response.}
    \centering
    \begin{tabular}{lp{4.6cm}|p{5.3cm}|c}
        & Role-play dialogue & GPT-4 & Score \\ \hline \hline
        CLI: & I don't have many friends, so she's the one who's been with me at school  & ~ & ~ \\ \hline
        COU: & I see. & [showing empathy] I see, your relationship with her is very important to you. [expresses understanding of the client's feelings] It's very painful that your relationship with her has turned out this way. & 1.00 \\ \hline
        CLI: & Yes.   & ~ & ~ \\ \hline
        COU: & From what you have told me so far, it seems to me that you feel text messages from her are bothersome and that you don't feel you can talk about anything with her,  & [to see what the client wants to do] What do you most want to do now? & 0.33 \\ \hline
        COU: & So I asked you if she is what you call a ``friend". & ~ & ~ \\ \hline
        CLI: & I feel lonely being alone at school. Don't you think so? & ~ & ~ \\ \hline
        COU: & I wonder... Which do you think is better, forcing yourself to be with someone you don't like very much, or being alone without forcing yourself, even though you might be lonely?
& [to empathize with the client] Sure, it's lonely to be alone at school, but it's important to create an environment where you feel comfortable. [to ascertain what the client would like to do] What would you like to do? & 1.66 \\ \hline
        % クライアント & 授業も集中できなくて、半分寝てます。だから宿題もわからなくて 受験勉強しなきゃいけないけど何をやればいいかもわからないです。 & ~ & ~ \\ \hline
        % カウンセラー & そうなんですね、、悪循環になってしまいますよね。 & [クライアントの話を受け入れるための受容] そうなんですね。授業が集中できないと、宿題もわからなくなってしまいますよね。受験勉強も何をやればいいかわからないというのは、とても大変そうですね。& 0.33 \\ \hline
        % クライアント & はい & ~ & ~ \\ \hline
        % カウンセラー & ちなみに、学校以外の時間はどんなことをして過ごしていることが多いですか？ & [クライアントが自分で解決策を見つけられるように、問題解決の手助けをするため]それは大変ですね。でも、一つずつ解決していけば、少しずつでも前に進めると思います。まずは、授業に集中できないことから始めてみてはどうでしょうか？何か授業に集中できない原因があるとしたら、何だと思いますか？ & 1.00 \\ \hline
    \end{tabular}
    
    \label{tb:dlg_gpt}
\end{table}

\section{Analysis}
\subsection{Subjective evaluation by professional counsellors}

To analyze the role-play dialogues and the GPT-4-generated utterances, third-party counsellors ranked both the utterances of the counsellors made during the role-play dialogues and the GPT-4-generated utterances. 
The utterances were rated according to a 3-point Likert scale from 0 (bad) to 2 (good), with three professional counsellors per dialogue. 
The counsellors also provided written explanations for their evaluations. A total of seven counsellors conducted evaluations. 
Table \ref{tb:dlg_gpt} shows examples of generated utterances and the average scores for each of the three counsellors' utterances.

We calculated Krippendorff's alpha for the dialogues of themes one through three (counselor utterances: 157; GPT-4 generated utterances: 124) in order to assess the agreement between the counselors' ratings of the utterances. The Krippendorff's alpha was 0.24, indicating a weak correlation.

\begin{figure}
\sidecaption
\includegraphics[scale=0.15]{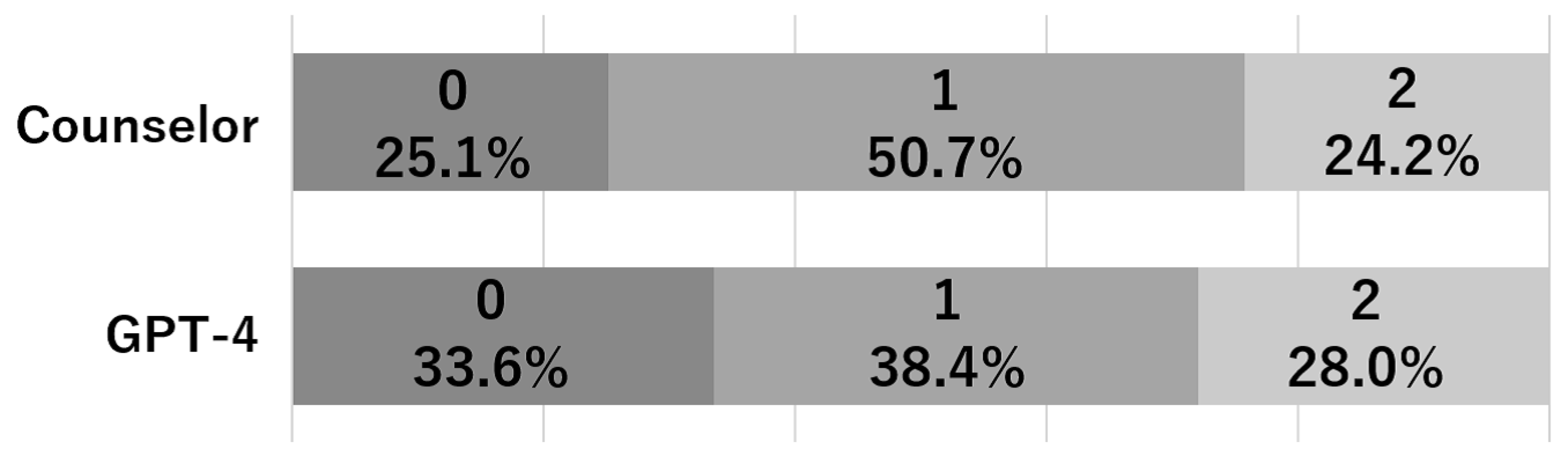}
\caption{Ratio of rating scores assigned to utterances}
\label{fig:retio}
\end{figure}

The mean rating scores for the counsellors' utterances and GPT-4-generated utterances were 0.99 (variance: 0.49) and 0.94 (variance: 0.61), respectively. 
We also conducted a Mann-Whitney U-test at a significance level of 0.05 and found no significant differences. 
The above results indicate that the difference in response quality between counsellors and GPT-4 is minimal. 
Figure \ref{fig:retio} shows the ratios of the ratings given to utterances by counsellors and GPT-4, respectively. 
As shown, a larger ratio of GPT-4 utterances was assigned a rating of 0 and 2 compared to the counsellor's utterances. 
More than half of the counsellors' utterances were given a score of 1; however, this was because the counsellors' utterances often included short utterances that were given a score of 1, such as ``I see" and ``Yes."

The evaluation results confirm that there were individual differences in the scoring tendencies of the evaluators. 
Therefore, we analyzed whether the same evaluator gave higher scores to counsellors' utterances or GPT-4-generated utterances when the context was identical. In cases where the counsellor made multiple consecutive utterances before the client responded, the average rating for all utterances was used as the rating for the counsellor's utterance.

The results of the comparison are shown in Table \ref{tb:comp}.
We found that more counsellors' utterances were rated as superior compared to GPT-4 utterances that were rated superior (34.8 \% vs 30.5\%).
However, the difference was small, and 34.7\% of the utterances were rated as equal.
We also confirmed that the quality of GPT-4 responses was very close to that of the human counsellors' responses.

\begin{table}[!htb]
    \centering
    \begin{tabular}{l|l} \hline
    Counsellor win & 284 (34.8\%) \\ 
    GPT-4 win  & 249 (30.5\%) \\
    Tie  & 283 (34.7\%) \\  \hline
    \end{tabular}
    \caption{Comparison of ratings for responses to the same context}
    \label{tb:comp}
\end{table}

We confirmed that LLM-generated responses were competitive with human responses despite having not used few-shot examples except for given annotated context in our prompt for GPT-4. Given that it is likely possible to further enhance the performance of GPT-4, we believe that there is potential for providing real-world counselling using an LLM-based system.

\subsection{Case Study}

When providing real-world counselling using a dialogue system, it is necessary to minimize inappropriate responses as much as possible. In this section, we analyze the responses generated by GPT-4 that were given low ratings.

The majority of responses with low ratings were phrases with inappropriate or unnatural phrasing and wording. An example of such an utterance is shown in Table \ref{tb:dlg_ex2}.
The counsellor who conducted the evaluation commented that the use of the word ``interesting" could cause discomfort for the client because it could imply that the questions were being asked out of curiosity.

\begin{table}
    \caption{Example of a generated response with a low rating due to use of inappropriate wording (Theme 1)}
    \centering
    \begin{tabular}{p{1.5cm}|l p{8.5cm}} \hline
        Context & Councelor: &  Okay, I see.\\ 
        & Client: &  I'm also fine when the person before me is performing.\\ \hline
        Generated response & GPT-4: &  That is interesting. So do you mean that you get more nervous right before your performance actually starts? \\ \hline
    \end{tabular}

    \label{tb:dlg_ex2}
\end{table}

There were also many responses where the GPT-4 response discusses the client's problem as if it were somebody else's problem. 
An example of such a case is shown in Table \ref{tb:dlg_ex3}. 
In this example, two of the three evaluators commented that the response sounds insincere, and one commented that the expression ``It looks difficult" should be avoided.
\begin{table}
    \caption{Example of generated response with low rating because it treated the client's problem as if it were somebody else's problem (Theme 4)}
    \label{tb:dlg_ex3}
    \centering
    \begin{tabular}{p{1.5cm}|l p{8.5cm}} \hline
        Context & Counsellor: &  
        If you don't mind, can you tell me when you feel like you are not studying as well as you would like?
        (You know what you have to do, but you don't have the motivation to do it. Or maybe you have a subject that you are not good at.)       \\ 
        & Client: & I can't concentrate in class, and I usually fall asleep. That's why I can't do my homework... I have to study for the entrance exam, but I don't even know what to do.\\ \hline
        Generated response& GPT-4: & I see. You won't be able to do your homework because you can't concentrate in class. It looks difficult that you don't know what to do with your exam studies. \\ \hline
    \end{tabular}
\end{table}

In particular, counsellors should take great care to avoid risky responses. 
A known problem with LLMs is that they can generate sentences with offensive or discriminatory content \cite{zhuo2023red, deshpande2023toxicity}.
The evaluators and the authors confirmed that no offensive or discriminatory statements were found among the GPT-4-generated utterances.
However, a small number of risky utterances were identified. An example of such an utterance is shown in Table \ref{tb:dlg_ex_risks22}. 
In response to this GPT-4-generated utterance, the evaluator wrote, ``The response that kindness causes one to suffer oneself runs the risk of inculcating the value that one should not be kind."

\begin{table}
    \caption{Example of generated risky utterance (Theme 2)}
    \centering
    \begin{tabular}{p{1.5cm}|l p{8.5cm}} \hline
        Context & Counsellor: & I see. You were thinking of her and putting her first. That is your kindness, and she also made you that way. \\ 
        & Client: &Ah, yes. I've been told for some time that I'm too kind.\\ \hline
        Generated response & GPT-4: & You are a kind person. But it is also true that your kindness has caused you to suffer. \\ \hline
    \end{tabular}
    \label{tb:dlg_ex_risks22}
\end{table}

Although the number of risky responses identified in this study was small, when input prompts include offensive content, LLMs tend to generate offensive sentences in response \cite{kumar2023certifying}.
In the role-play dialogues collected for this study, neither the counsellor nor GPT-4 generated responses included such content. In the real world, clients might provide responses with aggressive content. We plan to analyze such cases in future work.

\section{Conclusion}
We collected and annotated role-play counseling dialogue data, and professional counselors evaluated the appropriateness of GPT-4-generated responses. Analysis of the results demonstrated that the GPT-4-generated responses were competitive with those of human counselor responses. Furthermore, among the responses that received low ratings, no aggressive, discriminatory, or high-risk responses were identiﬁed.

In this study, we generated and evaluated responses based on the context of role-play dialogues. No experiments have demonstrated a dialogue system with a fully automatic counseling dialogue from start to ﬁnish. In future work, we plan to evaluate whether this system can provide fully automated, comprehensive counseling services.

\bibliographystyle{jplain}
\bibliography{main}

\end{document}